\documentclass[final,5p,times,twocolumn]{elsarticle}
\usepackage{amssymb}
\usepackage{amsmath}
\usepackage{hyperref}
\hypersetup{
    colorlinks=true, 
    citecolor=green,  
    linkcolor=red,  
    urlcolor=blue    
}
\usepackage{subcaption}
\usepackage{lipsum}
\usepackage{amsmath,amssymb,amsfonts}
\usepackage{mismath}
\usepackage{xcolor}
\usepackage[linesnumbered,ruled,vlined]{algorithm2e}
\usepackage{graphicx}
\usepackage{algpseudocode}
\usepackage{booktabs,caption}
\usepackage{textcomp}
\usepackage{tabularx}
\usepackage{xcolor}
\usepackage{amssymb}
\usepackage{pifont}
\newcommand{\cmark}{\ding{52}}%
\usepackage{colortbl}

\definecolor{gray1}{gray}{0.9}
\definecolor{gray2}{gray}{0.8}
\definecolor{gray3}{gray}{0.7}
\definecolor{lightgray}{gray}{0.92}
\usepackage{arydshln}

\usepackage{multirow}
\usepackage{algpseudocode}
\usepackage{hyperref} 
\usepackage{lineno,hyperref}
\usepackage{amsmath,amssymb,amsfonts}
\usepackage{graphicx}
\usepackage[utf8]{inputenc}
\usepackage[english]{babel}

\usepackage{indentfirst}
\usepackage{booktabs}
\usepackage{tabularx}
\usepackage[pscoord]{eso-pic}

\usepackage{amsmath}

\modulolinenumbers[5]
\journal{Knowledge-Based Systems}

\begin{document}

\begin{frontmatter}

\title{\textbf{Enhancing person re-identification via\\
Uncertainty Feature Fusion Method and Auto-weighted Measure
Combination} }

\author[Second,Third]{Quang-Huy Che}
\ead{huycq@uit.edu.vn}

\author[Second,Third]{Le-Chuong Nguyen}
\ead{21520655@gm.uit.edu.vn}

\author[Second,Third]{Duc-Tuan Luu}
\ead{tuanld@uit.edu.vn}

\author[Second,Third]{Vinh-Tiep Nguyen\corref{cor1}}
\ead{tiepnv@uit.edu.vn}

\cortext[cor1]{Corresponding author.}

\address[Second]{University of Information Technology, Ho Chi Minh City, Vietnam}
\address[Third]{Vietnam National University, Ho Chi Minh City, Vietnam.}
\begin{abstract}


Person re-identification (Re-ID) is a challenging task that involves identifying the same person across different camera views in surveillance systems. Current methods usually rely on features from single-camera views, which can be limiting when dealing with multiple cameras and challenges such as changing viewpoints and occlusions. In this paper, a new approach is introduced that enhances the capability of ReID models through the Uncertain Feature Fusion Method (UFFM) and Auto-weighted Measure Combination (AMC). UFFM generates multi-view features using features extracted independently from multiple images to mitigate view bias. However, relying only on similarity based on multi-view features is limited because these features ignore the details represented in single-view features. Therefore, we propose the AMC method to generate a more robust similarity measure by combining various measures. Our method significantly improves Rank@1 accuracy and Mean Average Precision (mAP) when evaluated on person re-identification datasets. Combined with the BoT Baseline on challenging datasets, we achieve impressive results, with a 7.9\% improvement in Rank@1 and a 12.1\% improvement in mAP on the MSMT17 dataset. On the Occluded-DukeMTMC dataset, our method increases Rank@1 by 22.0\% and mAP by 18.4\%. Code is available: \url{https://github.com/chequanghuy/Enhancing-Person-Re-Identification-via-UFFM-and-AMC}

\end{abstract}


\end{frontmatter}
\section{Introduction}

Person re-identification (Re-ID) is a crucial task in computer vision that aims to accurately identify individuals across multiple camera views. This is essential for various applications such as security, surveillance, and traffic management. To perform a query, the similarity between the query image and all images in the gallery is calculated based on features extracted from a pre-trained model. Measures like Euclidean distance and Cosine similarity are used to rank results by identifying the candidates that are most similar to a given query. The main characteristic of Re-ID is the collection of images from multiple cameras, capturing a person from different views, leading to significant differences within the intra-class. However, this task faces several challenges, such as cluttered backgrounds, changing views, and occlusion. These factors can cause a person's information to change or be incomplete, making it inaccurate to compare the similarity of images captured from different views. Despite significant progress made through deep learning and large-scale public datasets, Re-ID still has many challenges to overcome.
\begin{figure}[htp]
  \centering
        \includegraphics[width=0.5\textwidth]{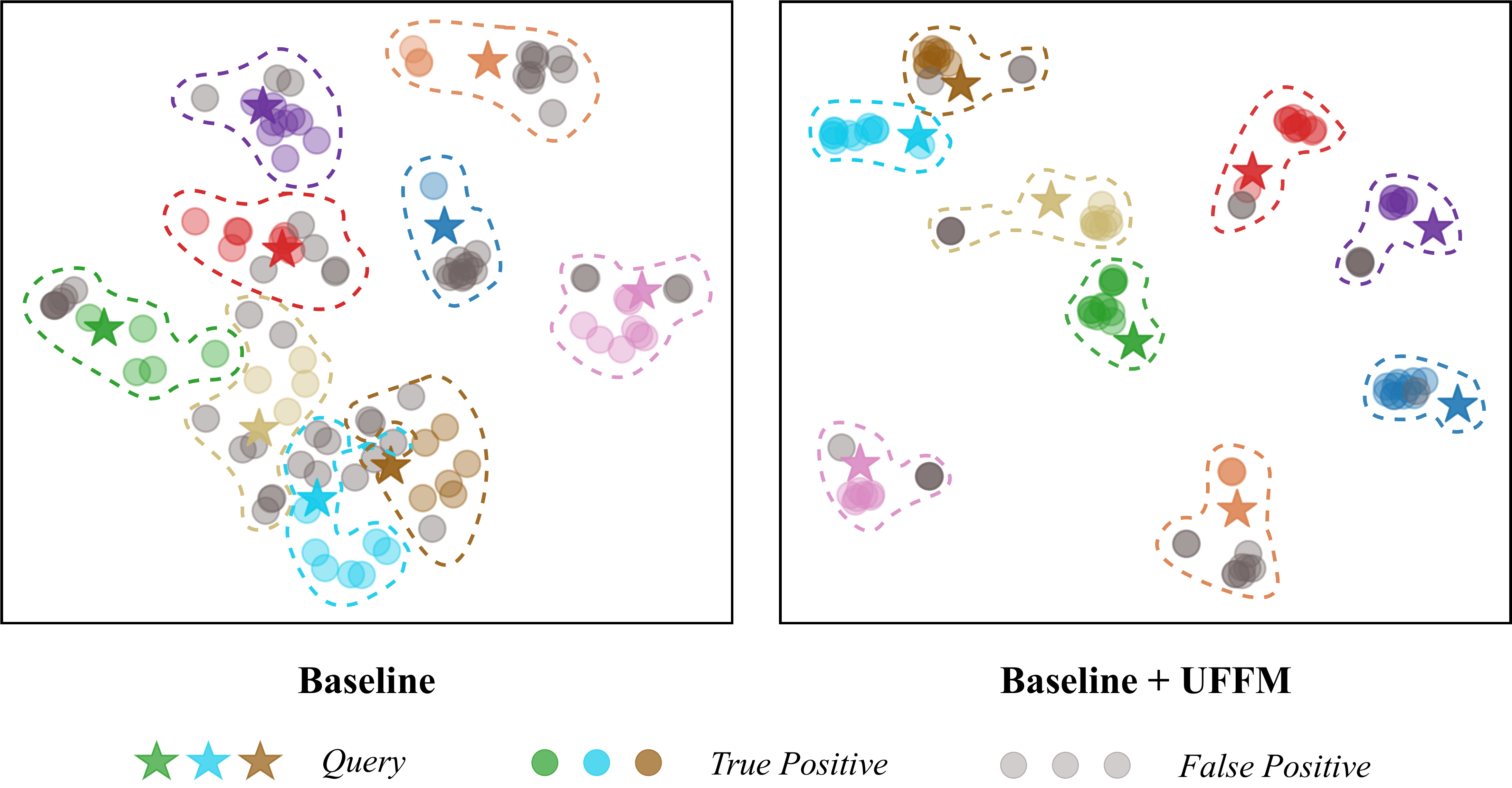}
        \caption{\textbf{t-SNE visualization:} Each cluster separates queries and their results from the others within a cluster comprising the top 20 query results. In a boundary, the star pentagon indicates the query feature, while dots of the same color as the petagrams represent correct query results and gray dots indicate incorrect results. The left figure shows the visual result of the Baseline, while the right figure shows the result of the Baseline combined with the proposed UFFM. The visual result demonstrates that when applying UFFM, the clusters tend to be separated, and the number of true positive points is greater than the number of false positive points in each cluster.}
        \label{tsne}
\end{figure}

\begin{figure*}[htp]
  \centering
        \includegraphics[width=0.8\textwidth]{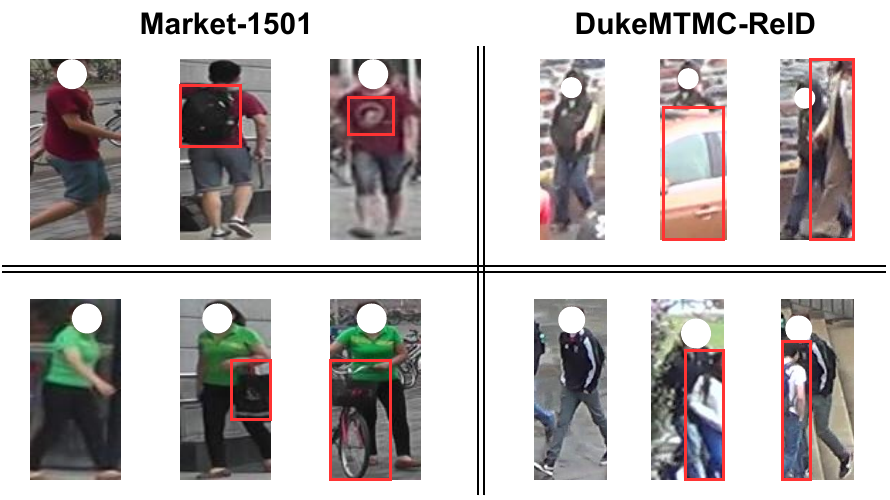}
        \caption{\textbf{Example of pedestrian images.} The Market-1501 and DukeMTMC-ReID datasets contain images of the same individuals from different camera views. In the examples from the Market1501 dataset, the \textcolor{red}{red rectangles} highlight information that appears in one frame but not in others. In contrast, in the DukeMTMC-ReID dataset, these \textcolor{red}{red boxes} refer to occluded regions where information about the person needs to be extracted.}
        \label{example}
\end{figure*}

Traditional ReID solutions rely on deep learning models trained to convert images into vector features \cite{PCB, SAN, TBE-Net, circle, FIDI, centroid, DAReID, Cross-modality, osfa, MGN+CFFS}. This training process ensures that pedestrian images from the intra-class have highly similar representations in the vector space. Recent works have optimized the learned feature embedding space and distance/similarity metrics using two methods: representation learning \cite{PCB, SAN, TBE-Net,BPBre,isp,hg,BoT,fastreid,clipreid,st-ReID,UMTS} and metric learning \cite{circle,FIDI,centroid,triplet,QAConv,OfM}. Previous methods extract separate features from images, which are computed independently and do not influence each other. In the context of multiple cameras, each feature of an image captured from a single viewpoint is considered a single-view feature. The single-view feature-based approach has clear weaknesses, particularly in multi-camera contexts. One of the main issues is the difference in observable regions of pedestrians from various viewpoints. This results in specific information about a person captured from one viewpoint not being present in another, even though both images depict the same individual. This inconsistency leads to significant differences between features obtained from different views, causing individuals within the same class to exhibit low similarity in scenarios where the camera viewpoint changes, or the person is partially occluded. This greatly reduces the effectiveness of single-view feature-based recognition methods, as important information may be lost or not fully considered. Specifically, Figure \ref{example} illustrates how the information about the same person may not always be fully visible in an image captured from a fixed view; for instance, a backpack may not be visible if the image is captured from the front view. Furthermore, when captured at different times, a person may be engaged in various activities and carrying different items, such as walking, holding a handbag, or riding a bicycle. Additionally, pedestrians may encounter challenges due to obstructions caused by objects or other individuals. Therefore, relying on a single image to represent an individual is insufficient. Single-view features are not reliable when images are taken from multiple cameras. Consequently, matching images captured from different viewpoints may lead to view bias. In person re-identification context, view bias refers to the tendency to misidentify individuals due to variations in camera perspectives. When a person is captured from different angles, significant changes in image features can occur, resulting in inaccurate image comparisons. How can we address view bias issue?

We propose addressing the limitations of single-view features by mitigating view bias by generating multi-view representations. However, it's important to note that the features in the embedding space are extracted independently, lacking the associated information needed for aggregation. Mikolaj et al. \cite{centroid} explored using pedestrian images with the same identity to create feature representations of a class by aggregating specific features across different camera views. This approach poses significant challenges in real-world scenarios, where images are collected from multiple cameras, making it difficult to accurately label each image to distinguish those with the same identity. Some studies with more complex computations \cite{MV-Reid,MultiviewAttention} propose generating 3D data for better generalization. However, these methods face difficulties reconstructing 3D images from 2D data, especially when the images contain occluded individuals or are of low quality. This makes it challenging for 3D models to recover the structure and details of the pedestrians. \cite{MFFN+MVMP, MVIIP} propose methods to train models that can aggregate features from multiple images into a single representative feature. \cite{MFFN+MVMP} trains the model to optimize the combination of single-view and multi-view features, helping mitigate view bias. Additionally, \cite{MVIIP} requires training to integrate information from multiple images of the same identity, creating a more comprehensive feature representation, especially in occlusion scenarios. However, both methods face the limitation of relying on a complex and time-consuming training process to achieve optimal results. Unlike previous works, we propose a method for aggregating single-view features without requiring class-level information or training the model. We introduce the Uncertainty Feature Fusion Method (UFFM) to address the limitations of previous approaches. UFFM aggregates features from multiple viewpoints by leveraging the nearest neighboring features of each image in the embedding space. Specifically, we identify neighboring features for each single-view feature and select the most similar ones to form multi-view features. Instead of aggregating by averaging neighboring features, UFFM applies weights based on the similarity between each neighboring feature and the single-view feature. This approach allows us to determine the contribution of each neighboring feature to the overall fusion process, resulting in a more comprehensive and informative representation of the object. Replacing single-view features with multi-view features not only allows for more comprehensive information capture about the persons but also mitigates bias that arises when only using features from a single viewpoint. Thus, UFFM significantly improves accuracy, especially in cases where the person is occluded or changes viewpoint. We conduct a thorough analysis using theoretical tools (Section \ref{propose}) and provide comments on the results (Section \ref{result}), as outlined below: (1) Multi-view features can be generated from single-view features in a straightforward manner. (2) Our method performs single-view feature selection in an unsupervised manner by considering the $K$ nearest neighbors. (3) UFFM can be seamlessly integrated with existing pre-trained models without requiring model training or fine-tuning. Utilizing multi-view features generated by the UFFM method helps mitigate view bias during the query. Figure \ref{tsne} simultaneously visualizes the features from query results based on both the conventional single-view features and our proposed multi-view features in a 2D coordinate space, which helps illustrate the effectiveness of our method. The query results based on single-view features, shown on the left, reveal features with high variance within each cluster corresponding to different query results. In contrast, the query results based on multi-view features, depicted on the right, show a more clearly separated distribution of results with reduced variance, leading to a more distinct feature distribution. This visualization demonstrates that multi-view features distinguish between classes with higher accuracy.

When observing a person from multiple viewpoints, we can gather comprehensive information about their appearance. This helps partially address the issue of view bias caused by factors such as changes in camera angles or occlusion. However, observations from different angles may lead to the loss of specific details, which are crucial for accurately distinguishing individuals. Similarly, while multi-view features help address the issue of view bias by generalizing a person's representation, they may also lose important details essential for precise individual differentiation. Based on this, each approach has advantages and disadvantages when comparing similarities using multi-view versus similarities using single-view features. This raises the question: Can combining both similarities leverage the detailed information from a single view while maintaining the comprehensive representation from multiple views? To address this issue, \cite{emd,vocvehicle,combine2008} have proposed combining different similarity measurements to maximize the strengths of each metric. Previous methods often relied on linear combinations of different measurements with manually selected weights. However, manually choosing weights results in a suboptimal solution, as it does not guarantee that the chosen weights yield good performance. Theoretically, we can maximize the strengths of different metrics by selecting good weights. So, how can we determine the ideal weights for combining measurements in the person re-identification task? In this study, we propose a robust method called Auto-weighted Measure Combination (AMC), which automatically determines the ideal weights to combine different similarity measurements between images in the person re-identification task. Instead of using fixed weights or manual selection, AMC adjusts the weights for each measure based on the specific objectives of the task: image features from the same class should have high similarity, while features from different classes should have low similarity. In this work, we propose AMC to combine the similarity between single-view features, multi-view features (generated by UFFM), and cross-camera encouragement (CCE \cite{cce}) to improve performance further. Using a linear regression algorithm, AMC learns optimal weights so that images of the same person are highly similar while images of different individuals are low. This approach eliminates manual experimentation with weight selection, which may not ensure optimal results. Instead, the combined weights are automatically calculated by AMC on the training dataset, providing objectivity in the testing phase. By doing so, AMC effectively leverages detailed information from single-view features and the comprehensiveness of multi-view features, thus enhancing accuracy in complex scenarios without manual tuning.

In summary, the contributions of this paper are as follows:
\begin{itemize}
    \item We discuss the issue of view bias in person re-identification within challenging multi-camera scenarios. To address this, we propose UFFM, which generates multi-view features from single-view features by selecting the nearest neighbor features unsupervised.
    
    \item To leverage different similarity measures, such as similarity based on multi-view feature similarity and single-view feature similarity, we introduce AMC. Our method automatically generates robust weights based on the task objectives, avoiding fixed or manual selection weights.
    
    \item Both proposed methods (UFFM and AMC) are implemented during the inference phase, requiring no training or fine-tuning of the person re-identification model. This allows for seamless implementation with pre-trained models.

    \item Our experiments, conducted on popular datasets like Market-1501 and DukeMTMC-ReID, as well as challenging datasets such as MSMT17 and Occluded-DukeMTMC, demonstrated the effectiveness of our method on key metrics such as Rank@1 and mAP for person re-identification tasks.
\end{itemize}

The rest of the article is presented as follows: We review related methodologies in Section \ref{relatedwork} to gain a better understanding of the advantages and limitations of Person ReID tasks. Our approach includes two main contributions: the Uncertainty Feature Fusion Method (UFFM) and Auto-weighted Measure Combination (AMC), which are detailed in Section \ref{propose}. We then proceed to evaluate our proposals on Person ReID datasets, including Market-1501, DukeMTMC-ReID, Occluded-DukeMTMC, and MSMT17 in Section \ref{result}. Finally, in Section \ref{conclusion}, we summarize our findings and propose potential avenues for future research.

\section{Related Work}\label{relatedwork}

Person re-identification (ReID) methods have been extensively researched using deep learning models. We present some popular approaches to deep learning-based person ReID and then demonstrate relevant measure combination methods associated with our proposal.

\subsection{Person Re-Identification}

In recent years, deep learning has made significant strides in person re-identification. Deep learning-based approaches can be divided into feature representation learning and deep metric learning. Feature representation learning focuses on automatically extracting features from images to enhance identification capabilities through various feature construction strategies. There are two main categories: local feature representation \cite{PCB, BPBre, isp, hg} and global feature representation \cite{BoT, fastreid, clipreid, st-ReID, UMTS}. Local feature representation emphasizes specific regions or parts of the person in the image, such as the head, shoulders, or legs, which helps in handling cases of occlusion or viewpoint changes. In contrast, global feature representation extracts information from the entire image, providing a more generalized and holistic representation. Some studies have also combined local and global features to create a more comprehensive and accurate representation, improving identification performance even in complex conditions \cite{SAN, TBE-Net}. On the other hand, deep metric learning focuses on optimizing the way the model compares two images to determine whether they belong to the same individual. These methods often involve the design of loss functions such as triplet loss \cite{triplet, centroid}, circle loss \cite{circle}, or contrastive loss \cite{contrastive}, which aim to optimize the distance between the features of the same individual while maximizing the distance between different individuals. Additionally, advanced sampling strategies are applied to effectively select pairs or groups of images for training, further enhancing model accuracy \cite{FIDI, QAConv, OfM}. Although extensively studied in single-view feature retrieval, the evaluation of the effectiveness of methods for multi-view features is still limited.

To address the challenge of view bias in single-view features, several multi-view feature extraction methods have been proposed \cite{centroid, MFFN+MVMP, face_mapr2023, MV-Reid, MultiviewAttention, MVIIP} to improve person re-identification performance. \cite{MV-Reid, MultiviewAttention} employ 3D multi-view image generation techniques to overcome viewpoint limitations in recognition tasks. However, these methods face significant challenges when reconstructing 3D data from 2D images, mainly when dealing with occlusions or low-quality images, making it difficult to recover the structure and finer details of the object. Additionally, 3D methods are computationally intensive, leading to increased costs and complexity compared to traditional 2D approaches Rather than relying on 3D image generation models. \cite{centroid, MFFN+MVMP, face_mapr2023, MVIIP} propose aggregating information from different images to create representative features. For instance, \cite{centroid,face_mapr2023} proposes aggregating the features of images within the same class by averaging them to find a representative feature. \cite{face_mapr2023} allows labeling features during face data collection, while \cite{centroid} uses class-level labels from person ReID datasets to aggregate features. In practice, assigning class labels to images collected from multiple views presents a significant challenge in person ReID, which limits the practical application of these methods. On the other hand, \cite{MFFN+MVMP} proposes a method to generate multi-view features by aggregation features from several fixed views. This method utilizes Graph Neural Networks (GNNs) to identify neighboring features without requiring pedestrian image labels belonging to intra-class. Despite this advancement, these methods require the construction of a high-cost graph during the training and testing phases. Moreover, they incorporate information from the query person's camera to generate multi-view features. This violates the cross-view matching setup during the query process, making unfair comparisons with other methods strictly following cross-view matching setting. Furthermore, Dong et al. \cite{MVIIP} propose leveraging information from multiple images of the same individual to create a comprehensive representation, improving recognition performance in scenarios where body parts are occluded. However, this method faces the challenge of generalizing to unseen situations during inference. If the model has not encountered images with similar characteristics during training, its recognition performance may decline during inference.

In this work, we introduce a novel method for generating multi-view features from single-view features without needing class-level labels. Unlike existing approaches that rely on heavy supervision or require significant intervention during the training phase, our method operates entirely during the query process, making it highly compatible with pre-trained models. This eliminates additional training or fine-tuning, reduces computational costs, and enables combining with pre-trained models. Additionally, our approach follows the cross-view matching setup, enabling fair comparison with other re-id methods.

\subsection{Measure Combination}

When solving retrieval problems, combining multiple measures can lead to better performance than using each measure individually. In a study \cite{combine2008}, similarities in a content-based image retrieval (CBIR) system are combined using independent similarity functions and a probability-based aggregate similarity function. The aim is to better reflect the semantic similarity between images by aggregating information from multiple sources. \cite{emdface} proposes combining cosine distance and Earth Mover's Distance (EMD) for face recognition. The authors use a two-stage process, where the cosine distance sorted the face list in the gallery in the first stage, and the top candidates are re-ranked in the second stage using EMD distance. However, the authors found that using only the EMD distance in the second stage did not improve accuracy and actually reduced it. The authors linearly combined the EMD and the cosine distance using the $\alpha$ coefficient, and they found that the proposed distance gave better results than using only the EMD distance. Zhu et al. \cite{vocvehicle} indicate that the similarity of two images can be biased due to similar backgrounds and shapes. To address this issue, the authors propose combining distances between images with shape and background similarities to reduce their influence. Previous methods of generating combined similarity/distance relied on manually selecting linear combination weights. However, this manual weight selection has limitations as it involves extensive trial and error, and the resulting weights are not guaranteed optimal. This approach leads to inefficiencies in performance, especially when the chosen weights fail to capture the combined measures' potential fully. Q. Huy et al. \cite{face_mapr2023} propose to combine cosine similarity with instance-based and centroid-based \cite{centroid} features in the face recognition task. Instead of manually selecting combined weights, they suggest using triplet feature algorithm and machine learning models to generate them. Although this approach shows promising initial results, this research does not explore the possibility of combining more than two different measures, especially in the case of person ReID, which requires compliance with cross-view matching settings.

This paper demonstrates the ability to combine multiple similarity measures in person re-identification by introducing Auto-weighted Measure Combination (AMC). Unlike previous approaches, AMC automatically generates combination weights without manual adjustment or selection, using the Triplet Data Generation Algorithm \ref{alg1}. The weight generation process is driven by the objective of the re-identification task, where features from the same individual exhibit high similarity, while those from different individuals display low similarity. Since the weight generation is automated, our method proves particularly effective when combining multiple measures. Furthermore, the weight generation occurs exclusively on the training set and is independent of the test set, thus ensuring objectivity during testing.

\begin{figure*}[]
\centering
\includegraphics[width=1.0\linewidth]{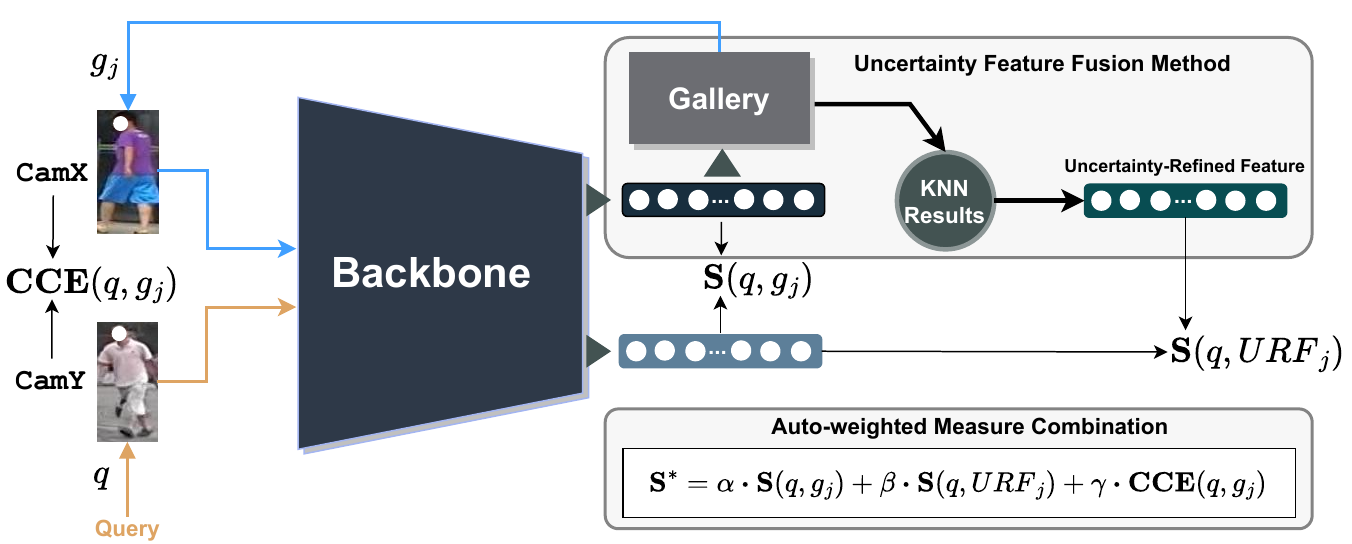}
\caption{\textbf{Overview of our proposed pipeline:} To determine the similarity between $q$ and $g_j$, we need to compute $\textbf{CCE}{(q, g_j)}$, $\textbf{S}{(q, g_j)}$, $\textbf{S}{(q, URF_j)}$ to generate the final similarity $\textbf{S}^*$. Where $q$ represents the query image and $g_j \in \mathcal{G}$ represents the $j^{th}$ image in the gallery. With the camera information for both images, we can easily compute $\textbf{CCE}{(q, g_j)}$. The shared backbone processes $q$ and $g_j$ for feature extraction. We directly compute $\textbf{S}{(q, g_j)}$ via the similarity of $q$ and single-view feature of $g_j$. Furthermore, we compute the similarity between $g_j$ and the images in the gallery to find the $K$ nearest similar images to $g_j$. Using weighted fusion, we obtain $f^{URF}_j$, from which we can easily compute $\textbf{S}{(q, URF_j)}$. Finally, the AMC method combines different measures and produces the robust similarity between $q$ and $g_j$ as $\textbf{S}^*$.}    
\label{pipeline}
\end{figure*}

\section{Proposed Method}\label{propose}

Our proposal consists of two main components: Uncertain Feature Fusion Method (UFFM) and Auto-weighted Measure Combination (AMC). In Figure \ref{pipeline}, you can see how these components are integrated into the re-identification model in the retrieval phase. When querying a person, features in the gallery are extracted independently, and their similarity to the query feature is calculated. Due to the limitations of single-view features, we propose to use UFFM to generate multi-view features that represent the library features. The representative feature, obtained through the UFFM proposal, is called \textbf{U}ncertainty-\textbf{R}e\textbf{F}ined (URF) feature. Now the similarity between $q$ and $g_j$ is denoted by \textbf{S}${(q, {URF}_j)}$ instead of \textbf{S}${(q, g_j)}$, where $q$ represents the query image and $g_j \in \mathcal{G}$ represents the j-$th$ image in the gallery. The URF feature of $g_j$ ($URF_j$) is not extracted directly from the model but is generated from the single-view features nearest to $g_j$.

We propose a method to automatically generate combination weights. Our robust measure is the combination of $\textbf{S}{(q, g_j)}$, $\textbf{S}{(q, URF_j)}$, $\textbf{CCE}{(q, g_j)}$. Where Cross-camera Encouragement term (CCE) \cite{cce} is responsible for promoting the trustworthiness of images from different cameras. With the AMC method, we can easily find the weights for the measure combination.

Next, the Uncertainty Feature Fusion Method (UFFM) is discussed in more detail in Section \ref{uffm}. Section \ref{wda} focuses on explaining the Auto-weighted Measure Combination (AMC).

\subsection{Uncertainty Feature Fusion Method (UFFM)}\label{uffm}

Directly generating features with the same identifier into a new composite feature poses challenges because it requires additional labels for images taken from multiple cameras. Our Uncertainty Feature Fusion Method (UFFM) addresses this by generating multi-view features from single-view features without the need for additional labels. Specifically, our strategy leverages the distribution in the learned embedding space to effectively aggregate features. Under the assumption that the model is well-trained, samples with identical identities naturally cluster together, while samples with different identities are separated. This approach, the Uncertainty Feature Fusion Method (UFFM), aligns with scientific principles to enhance clarity and accuracy in feature representation.

\begin{figure}[htp]
  \centering
        \includegraphics[width=0.8\linewidth]{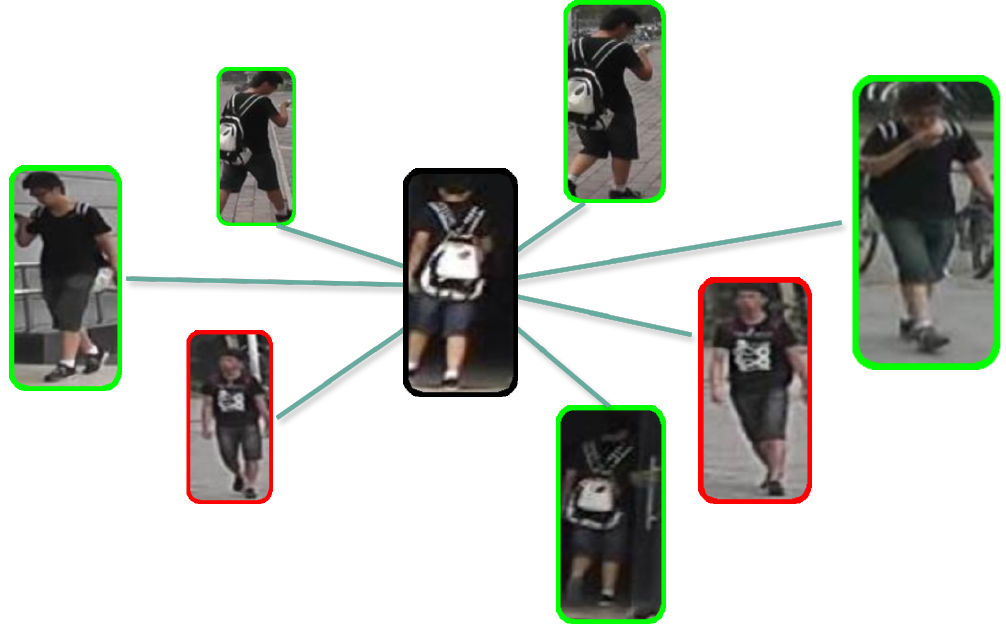}
        \caption{\textbf{An example of finding K nearest neighbors based on features, with K=7.} The black box represents the query object, the green boxes indicate correct results and the red boxes denote incorrect results.}
        \label{graph}
\end{figure}

We use a feature extractor $E_{\Theta}$, which is trained on the dataset $D_{\text{train}}$, to map an image $I \in \mathbb{R}^{3 \times H \times W}$ to a vector $f \in \mathbb{R}^{d}$, denoted as $f = E_{\Theta}(I)$. For each $f_j$ corresponding to a feature of $g_j$ in the Gallery set, we identify the set of $K$-nearest neighboring features from the Gallery set for each $f_j$, denoted by $F^{(\text{nn})}_j = \{f_{jk}^{(\text{nn})}\}_{k=1}^{K}$, where each $f_{jk}^{(\text{nn})} \in \mathbb{R}^{d}$. Although the selection of $K$ features nearest to $f_j$ is unsupervised, a well-structured embedding space still displays neighboring instances in the same class as $f_j$. However, incorrect results could introduce noise in the multi-view features, while more accurate results may enhance information accuracy. In Figure \ref{graph}, for each $f_j$, the multi-view feature is generated by considering the $K$ nearest neighbor features. This approach highlights features with the same identifier as $f_j$ in terms of quantity and similarity, allowing the generated multi-view feature to contain more information about features with the same identifier when generating the multi-view feature of $g_j$, denoted as $f^{URF}_j$. Directly averaging the features in the set $F_j^{(\text{nn})}$ is not always practical due to differences in similarity between $f_j$ and $f_{jk}^{(\text{nn})}$. Therefore, we use a weighted fusion approach, where similarity plays an important role in determining the contribution of each neighborhood to aggregate features. Specifically, the multi-view feature contains more information about highly similar features and less about features that are not very similar. This helps in removing noisy features during the unsupervised selection of $K$. This ensures that the contribution of each neighboring feature is proportionate to its similarity to $f$, facilitating a weighted fusion that captures the most relevant and significant features from the nearest neighbors.

\begin{equation}
f^{(\text{URF})}_j = \sum_{k=1}^{K} w_k \boldsymbol{\cdot} f_{jk}^{(\text{nn})}, \quad w_k = \frac{\text{cos}(f_j, f_{jk}^{(\text{nn})})}{\sum_{i=1}^{K} \text{cos}(f_j, f_{ji}^{(\text{nn})})}
\end{equation}

We prefer to use Cosine similarity over Euclidean or Cosine distance. Cosine similarity produces a value between -1 and +1. When calculating the weight, a negative value indicates that the similarity of $f_j$ and $f_{jk}^{(\text{nn})}$ is less than 0, meaning that the two vectors are opposite to each other. In this case, the negative weight enables $f^{(\text{URF})}_j$ to filter out the negative information. For consistency, we used cosine similarity in our work.

In the evaluation phase of most Re-Identification (ReID) problems, the cross-view matching setting is commonly applied. Initially, we define $\mathcal{Q}$ = $\{q_i\}_{i=1}^{N_q}$ is the query set and $\mathcal{G}$ = $\{g_j\}_{j=1}^{N_g}$ is the gallery set, where $N_q$ and $N_g$ are the number of images in the query set and the gallery set, respectively. Traditionally, for a query $q_i$, it is required to compute the similarities with each $g_j$ $\in$ $\mathcal{G}$ to obtain a similarity list $List$($q_i$, $\mathcal{G}$)=$\{\textbf{S}(q_i, g_j)\}_{j=1}^{N_g}$. Then, $List$($q_i$, $\mathcal{G}$) is sorted, excluding persons with both the same ID and camera from the result list. This implies that valid results are those captured by a different camera than that of the query. To ensure the cross-view matching setting in the feature aggregation process, we address this by selecting only the $K$ features captured by a different camera than that of the feature under consideration, or Cam$(f^{nn}_i)$ $\neq$ Cam$(f)$. This approach ensures that the fusion feature does not carry the characteristics of the query, thus maintaining fairness when using the cross-view matching setting in the evaluation process. After applying UFFM on each g$_j$ in the gallery set, resulting in $List$(q$_i$, $\mathcal{G}$)=$\{\textbf{S}(q_i, {URF}_j)\}_{j=1}^{N_g}$.

\subsection{Auto-weighted Measure Combination}\label{wda}

We propose to use multi-view features instead of single-view features in the query process through UFFM. However, the proposed multi-view features can generalize the person's information in the image, but this leads to the loss of detail of the features extracted from the image. On the other hand, the fixed-view features contain detailed information about the image. As a result, different similarity calculation methods produce different query results. With the goal of being able to combine different measures to find robust similarity, we introduce Auto-weighted Measure Combination (AMC) to determine linear combination weights. In this work, using AMC, we combined \textbf{S}${(q, g_j)}$ and \textbf{S}${(q, {URF}_j)}$, which represent the similarity of the query feature with the single-view feature and multi-view feature of a person in the gallery, respectively. Additionally, we combine the above similarities with the Cross-camera Encouragement term (CCE) \cite{cce} calculated as shown in Eq. \ref{cce}. In addition to increasing the reliability of two images taken from two different cameras in the cross-view matching setting, we aim to demonstrate the potential to simultaneously combine various measurements of the proposed AMC method.

\newcommand\mycommfont[1]{\footnotesize\ttfamily\textcolor{blue}{#1}}
\SetCommentSty{mycommfont}

\begin{algorithm}
\caption{Triplet Data Generation Algorithm}\label{alg1}

\SetKwInput{Input}{Input}
\SetKwInput{Output}{Output}

\SetKwFunction{FSimilarity}{cos}
\SetKwFunction{FCCE}{CCE}
\SetKwFunction{FCFFM}{CFFM}

\DontPrintSemicolon

\Input{$n \geq 0$ \tcc*{\fontfamily{qcr}\selectfont Number of samples}}
\Output{$D_{\text{triplet}}$}

$D_{\text{triplet}} \leftarrow \emptyset$\tcc*{\fontfamily{qcr}\selectfont Init training set for linear regression}
{\tcc {\fontfamily{qcr}\selectfont Loop to prepare data}}
\For{$n$ steps}{
  {\tcc {\fontfamily{qcr}\selectfont Extract feature step}}
  $x_a, x_p, x_n \leftarrow D_{\text{train}}$\;
  
  $f_a,f_p,f_n \leftarrow {E}_{\Theta}(x_a, x_p, x_n)$\;
  \BlankLine
  {\tcc {\fontfamily{qcr}\selectfont Prepare positive data}}
  $f^{RF}_p \leftarrow \FCFFM(f_p)$\;
  
  $\textbf{S}{(q, g_j) }_p \leftarrow \FSimilarity(f_a,f_p)$\;

  $\textbf{S}{(q, URF_j)}_p \leftarrow \FSimilarity(f_a,f^{RF}_p)$\;
  
  $\textbf{CCE}{(q, g_j)}_p \leftarrow \FCCE(x_a, x_p)$\;
  \BlankLine
  {\tcc {\fontfamily{qcr}\selectfont Prepare negative data}}
  $f^{RF}_n \leftarrow \FCFFM(f_n)$\;
  
  $\textbf{S}{(q, g_j) }_n \leftarrow \FSimilarity(f_a,f_n)$\;
  
  $\textbf{S}{(q, URF_j)}_n \leftarrow \FSimilarity(f_a,f^{RF}_n)$\;
  
  $\textbf{CCE}{(q, g_j)}_n \leftarrow \FCCE(x_a, x_n)$\;
  \BlankLine
    {\tcc {\fontfamily{qcr}\selectfont Put the data into the training set}}
  $D_{\text{triplet}} \leftarrow D_{\text{triplet}} \cup \bigg\{\Big[\textbf{S}{(q, g_j) }_p, \textbf{S}{(q, URF_j)}_p, \textbf{CCE}{(q, g_j)}_p \Big],  +1\bigg\}$\;
  
  $D_{\text{triplet}} \leftarrow D_{\text{triplet}} \cup \bigg\{\Big[\textbf{S}{(q, g_j) }_n, \textbf{S}{(q, URF_j)}_n, \textbf{CCE}{(q, g_j)}_n \Big], -1\bigg\}$\;
}
\end{algorithm}

\begin{equation}
\label{cce}
\begin{split}
\textbf{CCE}(q, g_j)=\begin{cases}
	1, & \text{Cam(q) = Cam($g_j$)}\\
        0, & \text{Cam(q) $\neq$ Cam($g_j$)}\\
		 \end{cases}
\end{split}
\end{equation} 

The combination measures include: $\textbf{S}{( q, g_j)}$, $\textbf{S}{(q, URF_j)}$, $\textbf{CCE}{(q, g_j)}$, and weights $\alpha$, $\beta$, $\gamma$.  
We can express the combination similarity equation as:

\begin{equation}
\label{fusion2}
\textbf{S}^* = \alpha \cdot \textbf{S}{(q, g_j) } + \beta \cdot \textbf{S}{(q, URF_j)} + \gamma \cdot \textbf{CCE}{(q, g_j)}
\end{equation}

\begin{table*}[]
\centering
\caption{\textbf{Comparison of our method with person ReID SOTA methods.} The best and second-best results are marked in \textbf{bold} and \underline{underline}, respectively.}
\label{sota_person}
\resizebox{\textwidth}{!}{%
\begin{tabular}{lcccccccc}
\toprule
\multicolumn{1}{c}{\multirow{2}{*}{\textbf{Methods}}}                  & \multicolumn{2}{c}{\textbf{Market-1501}} & \multicolumn{2}{c}{\textbf{DukeMTMC-ReID}} & \multicolumn{2}{c}{\textbf{MSMT17}} & \multicolumn{2}{c}{\textbf{Occluded-Duke}} \\ \cmidrule{2-9} 
\multicolumn{1}{c}{}                                                   & \textbf{Rank@1}     & \textbf{mAP}       & \textbf{Rank@1}      & \textbf{mAP}        & \textbf{Rank@1}   & \textbf{mAP}    & \textbf{Rank@1}       & \textbf{mAP}       \\ \midrule
BoT Baseline \cite{BoT} \textcolor{gray}                   & 94.5                & 85.9               & 86.4                 & 76.4                & 74.1              & 50.2            & 48.6                  & 42.6               \\
CLIP-ReID Baseline \cite{clipreid}\textcolor{gray} & 94.7                & 88.1               & 88.6                 & 79.3                & 82.1              & 60.7            & 54.2                  & 47.4               \\ \midrule
ISP \cite{isp} \textcolor{gray}{\small\textit{\fontfamily{pcr}\selectfont{(ICCV’20)}}}                    & -                   & -                  & 88.7                 & 78.9                & -                 & -               &          62.8    &   52.3   \\
RGA-SC \cite{RGASC} \textcolor{gray}{\small\textit{\fontfamily{pcr}\selectfont{(CVPR’20)}}}                    & \underline{96.1}     &      88.4               & 86.9    & 75.6                & 80.3 
    &      57.5               &          -    &   -   \\
HG \cite{hg} \textcolor{gray}{\small\textit{\fontfamily{pcr}\selectfont{(BMVC’21)}}}                      & 95.6                & 86.1               & 87.1                 & 77.5                & -                 & -               &                       &                    \\
PAT \cite{pat}\textcolor{gray}{\small\textit{\fontfamily{pcr}\selectfont{(CVPR’21)}}}                     & 95.4                & 88.0               & 88.8                 & 78.2                & -                 & -                   & 64.5    &                53.6      \\
MGN+CFFS \cite{MGN+CFFS}\textcolor{gray}{\small\textit{\fontfamily{pcr}\selectfont{(KBS’21)}}}                  & 94.8                & 87.5               & 89.9             & 79.9                & -              & -            &    -               &      -           \\
DAReID  \cite{DAReID}\textcolor{gray}{\small\textit{\fontfamily{pcr}\selectfont{(KBS’21)}}}                  & 94.6                & 87.0               & 88.9             & 78.4                & -              & -            &  -                   &    -           \\
CAL \cite{cal}\textcolor{gray}{\small\textit{\fontfamily{pcr}\selectfont{(ICCV’21)}}}                     & 94.5                & 87.0               & 87.2                 & 76.4                & 79.5              & 56.2            &                       &                    \\
OfM \cite{OfM}\textcolor{gray}{\small\textit{\fontfamily{pcr}\selectfont{(AAAI’21)}}}                     & 94.9                & 87.9               & 89.0                 & 78.6                & -                 & -               &                       &                    \\
MGN+CircleLoss \cite{circle}\textcolor{gray}{\small\textit{\fontfamily{pcr}\selectfont{(AAAI’21)}}}       & \underline{96.1}      & 87.4               & -                    & -                   & 76.9              & 52.1            &                       &                    \\
DRL-Net \cite{drl} \textcolor{gray}{\small\textit{\fontfamily{pcr}\selectfont{(TMM’22)}}}                 & 94.7                & 86.9               & 88.1                 & 76.6                & 78.4              & 55.3     &  65.0        &            50.8          \\
QAConv-GS \cite{QAConv}\textcolor{gray}{\small\textit{\fontfamily{pcr}\selectfont{(CVPR’22)}}}            & 91.6                & 75.5               & -                    & -                   & 79.1              & 49.5            &                       &                    \\
CLIP-ReID \cite{clipreid}\textcolor{gray}{\small\textit{\fontfamily{pcr}\selectfont{(AAAI’22)}}}          & 95.7                & 89.8               & 90.0                 & \underline{84.7}      & \textbf{84.4}     & \underline{63.0}                  & 61.0    & 53.5         \\
AGW \cite{AGW}\textcolor{gray}{\small\textit{\fontfamily{pcr}\selectfont{(TPAMI’22)}}}                    & 95.1                & 87.8               & 89.0                 & 79.6                & 68.3              & 49.3            &                       &                    \\
FastReID \cite{fastreid}\textcolor{gray}{\small\textit{\fontfamily{pcr}\selectfont{(ACM MM’23)}}}         & 95.4                & 88.2               & 89.6                 & 79.8                & 83.3              & 59.9            &                       &                    \\
BPBreID \cite{BPBre}\textcolor{gray}{\small\textit{\fontfamily{pcr}\selectfont{(WACV’23)}}}               & 95.1                & 87.0               & 89.6                 & 78.3                & -                 & -               & 66.7                 & 54.1                   \\
MV\text{$I^2$}P \cite{MVIIP}\textcolor{gray}{\small\textit{\fontfamily{pcr}\selectfont{(Inf Fusion’24)}}}               & 95.2                & 87.0               & -                 & -                & 80.4                 & 56.4               & 68.2                 & 55.6                   \\

\hline
BoT Baseline + Ours                                                             & \textbf{96.2}       & \underline{91.0}     & \underline{90.3}       & 83.1                & 82.0              & 62.3            & \textbf{70.6}                  & \underline{61.0}               \\
CLIP-ReID Baseline + Ours                                               & \underline{96.1}      & \textbf{92.0}      & \textbf{91.3}        & \textbf{85.0}       & \underline{83.8}    & \textbf{67.6}   & \underline{68.9}                  & \textbf{61.9}               \\ \bottomrule
\end{tabular}
}
\end{table*}

We use the Linear Regression model to determine the weights $\alpha$, $\beta$, and $\gamma$. Essentially, $\textbf{S}^*$ represents the sum of the Cosine similarities. We expect $\textbf{S}^*$ to have values in the range of $[-1, +1]$, where $+1$ indicates perfect similarity and $-1$ indicates complete dissimilarity between two persons. Therefore, in our regression model, if we take the measurements of $q$ and $g_j$ as input, the labels are $+1$ if $q$ and $g_j$ have the same identifier and $-1$ otherwise. The goal of the linear regression model is to find the weights so that if we input measurements of two images with the same identifier, the model approaches $+1$, and if the two images have different identifiers, the model approaches $-1$. We create data for the linear regression model by randomly selecting $n$ sets of triples: anchor (x$_a$), positive (x$_p$), and negative (x$_n$), where x$_a$, x$_p$, and x$_n$ represent images in the dataset. For each pair (x$_a$, x$_p$), we generate a data point for the training set labeled as +1, and for (x$_a$, x$_n$) labeled as -1.


To be independent of the evaluation phase, we use the training dataset $\textbf{D}_{train}$ to generate data for the linear regression model. Unlike the evaluation dataset, the training dataset contains images with identifier information captured from different cameras. Instead of using UFFM to generate Uncertain Refined feature $f^{(\text{URF})}$ for computing $\textbf{S}{(q, URF_j) }$, we propose to generate Refined feature using a method called Certain Feature Fusion Method (CFFM). Where the refined feature, $f^{(\text{RF})}$ is described in Equation \ref{cmvf}. 

\begin{equation}
\label{cmvf}
f^{(\text{RF})} = \frac{1}{K}\sum_{k=1}^{K}  f^{(cnn)}_k
\end{equation}

Where $F^{cnn}$ = $\{f_k^{cnn}\}_{k=1}^{K}$ represents the set of Certain Nearest-Neighbor features that are most similar to $f$ and share the same identity as $f$. The outcomes in \ref{nn_sec} also show that using CFFM leads to better results compared to UFFM. Algorithm \ref{alg1} details the process for generating training data for a linear regression model using the Triplet Data Generation Algorithm. Following model training, the linear regression equation can be expressed as:

\begin{equation}
\label{fusion3}
y = w_3 \cdot \textbf{S}{(q, g_j) } + w_2 \cdot \textbf{S}{(q, RF_j)} + w_1 \cdot \textbf{CCE}{(q, g_j)} + w_0
\end{equation}

By assigning the values $w_3$, $w_2$, and $w_1$ to $\alpha$, $\beta$, and $\gamma$ and ignore $w_0$, we can utilize $S^*$ as the robust similarity for the query.

\section{Experimental Results}\label{result}

\subsection{Datasets and Settings}
\subsubsection{Datasets}
We conduct training and evaluation of the proposed model on four Person ReID datasets: Market1501\cite{market}, DukeMTMC-ReID\cite{duke} and MSMT17\cite{msmt17}, Occluded-DukeMTMC\cite{occluded}:
\begin{itemize}
    \item Market-1501\cite{market}: The Market-1501 dataset is collected using six cameras positioned in front of a supermarket at Tsinghua University. It comprises a total of 32,668 images belonging to 1,501 individuals. The dataset is divided into 12,936 training images with 751 individuals and 19,732 testing images with 750 individuals. Among the test images, 3,368 are selected as the query set.

    \item DukeMTMC-ReID\cite{duke}: The DukeMTMC-ReID dataset, a subset of DukeMTMC, is tailored for image-based person re-identification. Captured from high-resolution videos by eight cameras, it's a significant pedestrian image dataset. Images are meticulously cropped using hand-drawn bounding boxes. It comprises 16,522 training images covering 702 identities, 2,228 query images of the other 702 identities, and 17,661 gallery images.

    \item MSMT17\cite{msmt17}: MSMT17 is a comprehensive multi-scene multi-time person re-identification dataset. It includes 180 hours of video footage captured by 12 outdoor cameras and 3 indoor cameras across 12 different time slots. With 4,101 annotated identities and 126,441 bounding boxes, it offers extensive data for re-identification research.

    \item Occluded-DukeMTMC\cite{occluded}: The dataset includes 15,618 training images, 17,661 gallery images, and 2,210 occluded query images. Our experimental results on the Occluded-DukeMTMC dataset highlight the effectiveness of our method in addressing Occluded Person Re-ID challenges. Notably, our approach eliminates the need for any manual cropping during the preprocessing stage.

\end{itemize}

\begin{table*}[]
\centering
\caption{Evaluation of our proposed methods with BoT Baseline and CLIP-ReID Baseline}
\label{ab1}
\centering
\resizebox{1.05\textwidth}{!}{%
\begin{tabular}{lcccccccccc}
\toprule
\multirow{2}{*}{\textbf{Baseline}}                        & \multicolumn{1}{l}{\multirow{2}{*}{\textbf{UFFM}}} & \multicolumn{1}{l}{\multirow{2}{*}{\textbf{AMC}}} & \multicolumn{2}{c}{\textbf{Market-1501}}                                                                                                                                                                      & \multicolumn{2}{c}{\textbf{DukeMTMC-ReID}}                                                                                                                                                                     & \multicolumn{2}{c}{\textbf{MSMT17}}    & \multicolumn{2}{c}{\textbf{Occluded-DukeMTMC}}                                                                                                                                                                      \\ \cmidrule{4-11} 
                                             &                       &                      & \textbf{Rank@1}                                                                                             & \textbf{mAP}                                                                                               & \textbf{Rank@1}                                                                                             & \textbf{mAP}                                                                                                & \textbf{Rank@1}                                                                                             & \textbf{mAP}                   & \textbf{Rank@1}                                                                                             & \textbf{mAP}                                                                                                 \\ \hline
\multirow{5}{*}{\begin{tabular}[c]{@{}l@{}}BoT \\ Baseline\end{tabular}}            & \multicolumn{1}{l}{}                      &                      & 94.5                                                                                              & 85.9                                                                                              & 86.4                                                                                              & 76.4                                                                                               & 74.1                                                                                              & 50.2        & 48.6 &    42.6                                                                                  \\ \cdashline{2-11}
                                             & {\color[HTML]{3531FF} \cmark }            &                                           & \begin{tabular}[c]{@{}c@{}}96.1\\ {\color[HTML]{3531FF} $\uparrow$ 1.6}\end{tabular}              & \begin{tabular}[c]{@{}c@{}}91.2\\ {\color[HTML]{3531FF} $\uparrow$ 5.3}\end{tabular}              & \begin{tabular}[c]{@{}c@{}}90.2\\ {\color[HTML]{3531FF} $\uparrow$ 3.8}\end{tabular}              & \begin{tabular}[c]{@{}c@{}}83.3\\ {\color[HTML]{3531FF} $\uparrow$ 6.9}\end{tabular}               & \begin{tabular}[c]{@{}c@{}}81.5\\ {\color[HTML]{3531FF} $\uparrow$ 7.4}\end{tabular}              & \begin{tabular}[c]{@{}c@{}}61.8\\ {\color[HTML]{3531FF} $\uparrow$ 11.6}\end{tabular}  &
                                             \begin{tabular}[c]{@{}c@{}}71.1\\ {\color[HTML]{3531FF} $\uparrow$ 22.5}\end{tabular}&
                                             \begin{tabular}[c]{@{}c@{}}61.8 \\ {\color[HTML]{3531FF} $\uparrow$ 19.2}\end{tabular}
                                             \\ \cdashline{2-11} 
                                             & {\color[HTML]{3531FF} \cmark }            & {\color[HTML]{3531FF} \cmark }            & \begin{tabular}[c]{@{}c@{}}96.2$_{\pm0.05}$\\ {\color[HTML]{3531FF} $\uparrow$ 1.7}\end{tabular}  & \begin{tabular}[c]{@{}c@{}}91.0$_{\pm 0.1}$\\ {\color[HTML]{3531FF} $\uparrow$ 5.1}\end{tabular}  & \begin{tabular}[c]{@{}c@{}}90.3$_{\pm 0.09}$\\ {\color[HTML]{3531FF} $\uparrow$ 3.9}\end{tabular} & \begin{tabular}[c]{@{}c@{}}83.1$_{\pm 0.13}$\\ {\color[HTML]{3531FF} $\uparrow$ 6.7}\end{tabular}  & \begin{tabular}[c]{@{}c@{}}82.0$_{\pm 0.24}$\\ {\color[HTML]{3531FF} $\uparrow$ 7.9}\end{tabular} & \begin{tabular}[c]{@{}c@{}}62.3$_{\pm 0.3}$\\ {\color[HTML]{3531FF} $\uparrow$ 12.1}\end{tabular} &
                                             \begin{tabular}[c]{@{}c@{}}70.6$_{\pm 1.44}$\\ {\color[HTML]{3531FF} $\uparrow$ 22.0}\end{tabular}&
                                             \begin{tabular}[c]{@{}c@{}}61.0$_{\pm 0.83}$\\ {\color[HTML]{3531FF} $\uparrow$ 18.4}\end{tabular}
                                             \\ \midrule
\multirow{5}{*}{\begin{tabular}[c]{@{}l@{}}CLIP-ReID \\ Baseline\end{tabular}} &                                           &                                           & 94.7                                                                                              & 88.1                                                                                              & 88.6                                                                                              & 79.3                                                                                               & 82.1                                                                                              & 60.7    
                & 54.2 & 47.4
                \\ \cdashline{2-11}
                                             & {\color[HTML]{3531FF} \cmark }            &                                           & \begin{tabular}[c]{@{}c@{}}95.7\\ {\color[HTML]{3531FF} $\uparrow$ 1.0}\end{tabular}              & \begin{tabular}[c]{@{}c@{}}92.0\\ {\color[HTML]{3531FF} $\uparrow$ 3.9}\end{tabular}              & \begin{tabular}[c]{@{}c@{}}90.7\\ {\color[HTML]{3531FF} $\uparrow$ 2.1}\end{tabular}              & \begin{tabular}[c]{@{}c@{}}84.8\\ {\color[HTML]{3531FF} $\uparrow$ 5.5}\end{tabular}               & \begin{tabular}[c]{@{}c@{}}83.4\\ {\color[HTML]{3531FF} $\uparrow$ 1.3}\end{tabular}              & \begin{tabular}[c]{@{}c@{}}67.7\\ {\color[HTML]{3531FF} $\uparrow$ 7.0}\end{tabular}  
                                             & \begin{tabular}[c]{@{}c@{}}68.6\\ {\color[HTML]{3531FF} $\uparrow$ 14.4}\end{tabular}
                                             & \begin{tabular}[c]{@{}c@{}}62.1\\ {\color[HTML]{3531FF} $\uparrow$ 14.7}\end{tabular} 
                                             \\ \cdashline{2-11} 
                                             & {\color[HTML]{3531FF} \cmark }            & {\color[HTML]{3531FF} \cmark }            & \begin{tabular}[c]{@{}c@{}}96.1$_{\pm 0.13}$\\ {\color[HTML]{3531FF} $\uparrow$ 1.4}\end{tabular} & \begin{tabular}[c]{@{}c@{}}92.0$_{\pm 0.05}$\\ {\color[HTML]{3531FF} $\uparrow$ 3.9}\end{tabular} & \begin{tabular}[c]{@{}c@{}}91.3$_{\pm 0.45}$\\ {\color[HTML]{3531FF} $\uparrow$ 2.7}\end{tabular} & \begin{tabular}[c]{@{}c@{}}85.0 $_{\pm 0.18}$\\ {\color[HTML]{3531FF} $\uparrow$ 5.7}\end{tabular} & \begin{tabular}[c]{@{}c@{}}83.8$_{\pm 0.36}$\\ {\color[HTML]{3531FF} $\uparrow$ 1.7}\end{tabular} & \begin{tabular}[c]{@{}c@{}}67.6$_{\pm 0.55}$\\ {\color[HTML]{3531FF} $\uparrow$ 6.9}\end{tabular}   & \begin{tabular}[c]{@{}c@{}}68.9$_{\pm 1.37}$\\ {\color[HTML]{3531FF} $\uparrow$ 14.7}\end{tabular} 
                                             &\begin{tabular}[c]{@{}c@{}}61.9$_{\pm 0.9}$\\ {\color[HTML]{3531FF} $\uparrow$ 14.5}\end{tabular} \\
                                             \bottomrule
\end{tabular}%
}
\end{table*}

\subsubsection{Implementation Details}

In this study, we utilize BoT Baseline \cite{BoT} and CLIP-ReID Baseline \cite{clipreid} as Baselines, employing a ResNet50 architecture as its backbone. Our methodology leverages readily available pre-trained models, eliminating the necessity for retraining. We standardize the dimensions of all images to 256 $\times$ 128 for the inference process. We select the hyperparameters $K$ and $n$ through visualization and present the chosen values in Section \ref{hyper}. We aim to select hyperparameters that balance efficiency across all datasets rather than optimizing for each dataset individually. For the UFFM method, we set $K=4$ for Market-1501, DukeMTMC-ReID, Occluded-DukeMTMC, and $K=6$ for MSMT17. For the results obtained by integrating AMC, we set $n=400$ for all datasets. Additionally, we conduct the algorithm five times to ensure fairness due to the random nature of the Triplet Data Generation Algorithm. Consistent with previous works \cite{BoT,clipreid}, the evaluation metrics used are the Cumulative Matching Characteristics (CMC) at Rank@1 and the mean Average Precision (mAP). Rank@1 measures the average precision of the first result returned for each query image. The mAP represents the average precision value when the query results are sorted by similarity, the closer the correct result is to the top of the list, the higher the score.

\subsection{Comparison with Person ReID State-of-the-Art Methods}

In our study, we benchmark our method against other person ReID approaches, as detailed in Table \ref{sota_person}. This table provides an in-depth comparison of the performance of various methods across three standard datasets in person re-identification: Market-1501, DukeMTMC-ReID, and MSMT17. Our comparison primarily focuses on methods that employ a ResNet50 backbone and resize images to 256 $\times$ 128.

Integrating our approach with two Baselines, the BoT Baseline and CLIP-ReID Baseline, significantly improves performance compared to the Baseline alone, as shown in Table \ref{sota_person}. In the Occluded-DukeMTMC dataset, we observ a substantial 22.0\% increase in Rank@1 and an 18.4\% increase in mAP when combining the BoT Baseline with our proposed methods. Similarly, integrating the CLIP-ReID Baseline with our methods results in a 14.7\% increase in Rank@1 and a 14.5\% increase in mAP. Combining our method with Baselines yields the best results compared to the previous state of the art, showcasing our method's effectiveness and adaptability. When integrating our method with Baselines, we also witness significant performance improvements on the challenging MSMT17 dataset. Combining the CLIP-ReID Baseline with our method results in an mAP of 67.6, surpassing the best previous methods with a 4.3\% mAP increase. Our method's effectiveness is further demonstrated on the Market1501 and DukeMTMC-ReID datasets, highlighting its adaptability and scalability across diverse and complex datasets.

\subsection{Ablation study}

\subsubsection{Effectiveness of UFFM and AMC}

Table \ref{ab1} demonstrates the performance of our method when combined with the BoT Baseline \cite{BoT} and CLIP-ReID Baseline \cite{clipreid}. This table provides a detailed overview of the performance improvements our proposed methods offer over the baselines on four datasets. In the results, utilizing the proposed UFFM leads to significant enhancements in Rank@1 and mAP metrics. Furthermore, the addition of AMC demonstrates the capacity to improve the accuracy of the Rank@1 measure, but sometimes at the expense of reducing the mAP performance compared to when combined with UFFM in some cases. These improvements indicate that combining UFFM and AMC can enhance accuracy and bolster the model's generalization ability on large and complex datasets. This proves that these methods effectively improve re-identification performance under challenging conditions.

\begin{table}[]
\centering
\caption{Analysis of computation cost on Market1501 and DukeMTMC-ReID}
\label{cost}
\begin{tabular}{lcc}
\hline
\multicolumn{1}{c}{\textbf{}} & \textbf{Market-1501}   & \textbf{DukeMTMC-ReID} \\ \hline
\textbf{BoT Baseline}         & $1.9\times10^{-4}$ (s) & $1.5\times10^{-4}$ (s) \\ \hline
\textbf{+UFFM}                & $2.0\times10^{-1}$ (s) & $2.6\times10^{-1}$ (s) \\ \hline
\textbf{+UFFM+AMC}            & 1.2 (s)                & $7.8\times10^{-1}$ (s) \\ \hline
\end{tabular}%
\end{table}

\subsubsection{Analysis of Time Cost}


This section analyzes the computational time when incorporating our method to compute similarity (excluding image feature extraction and accuracy computation). The results are presented in Table \ref{cost}, showing that the combination of our proposed methods results in higher time cost due to the additional calculations. Our method's computational time is higher than calculations based on the original features; however, our method still ensures low latency during computation. This analysis demonstrates the trade-off between computational cost and accuracy in our method.


\subsubsection{Effect of UFFM and CFFM} \label{nn_sec}

Table \ref{uffm_and_cffm} shows the results of the Feature Fusion Methods in the Triplet Data Generation Algorithm \ref{alg1}. The results indicate that both Rank@1 and mAP are improved when using the CFFM method to aggregate features in the Triplet Data Generation Algorithm across both the Market-1501 and DukeMTMC-ReID datasets, with a pre-trained CLIP-ReID Baseline. This shows that the guaranteed multi-view feature aggregation based on the training set labels allows the Triplet Data Generation Algorithm \ref{alg1} to create more good weights.
\begin{table}[]
\caption{Comparison of Accuracy using UFFM and CFFM}
\label{uffm_and_cffm}
\centering
\begin{tabular}{lcccc}
\toprule
\multirow{2}{*}{} & \multicolumn{2}{c}{\textbf{Market-1501}}                   & \multicolumn{2}{c}{\textbf{DukeMTMC-ReID}}                 \\ \cmidrule{2-5} 
                  & \textbf{Rank@1} & \textbf{mAP} & \textbf{Rank@1} & \textbf{mAP} \\ \midrule
UFFM              & 95.7   & 91.9                         & 90.6   & 84.7                        \\ \hdashline
CFFM              & 96.1   & 92.0                         &  91.3  &  85.0                          \\ \bottomrule
\end{tabular}%
\end{table}

\begin{table}[]
\caption{Ablation study of measures using AMC on Market-1501.}
\label{ab_distance}
\centering

\begin{tabular}{ccccc}
\toprule
\multicolumn{3}{c}{\textbf{AMC}}                                                                        & \multicolumn{2}{c}{\textbf{Market-1501}} \\ \midrule
\textbf{\textbf{S}${(q, g_j)}$} & \textbf{\textbf{S}${(q, URF_j)}$} & \textbf{\textbf{CCE}${(q, g_j)}$} & \textbf{Rank@1}      & \textbf{mAP}      \\ \midrule
\color[HTML]{3531FF} \cmark     &                                   &                                   & 94.7                 & 88.1              \\
                                & \color[HTML]{3531FF} \cmark       &                                   & 95.7                 & 92.0              \\
\color[HTML]{3531FF} \cmark     & \color[HTML]{3531FF} \cmark       & \multicolumn{1}{l}{}              & 95.9                 & 91.9              \\
\color[HTML]{3531FF} \cmark     &                                   & \color[HTML]{3531FF} \cmark       & 95.4                 & 88.8              \\
\color[HTML]{3531FF} \cmark     & \color[HTML]{3531FF} \cmark       & \color[HTML]{3531FF} \cmark       & 96.1                 & 92.0              \\ \bottomrule
\end{tabular}
\end{table}

\subsubsection{Performance comparison combining different measures when using AMC}
Table \ref{ab_distance} illustrates the performance of the CLIP-ReID Baseline model. Using the Auto-weighted Measure Combination (AMC) with different measures on the Market-1501 dataset shows that combining measures consistently produced better results compared to using a single measure. Specifically, combining three measures yields the best results. Therefore, the experimental results indicate that our method effectively combines different measures.
\begin{figure}[htp]
  \centering
        \includegraphics[width=0.5\textwidth]{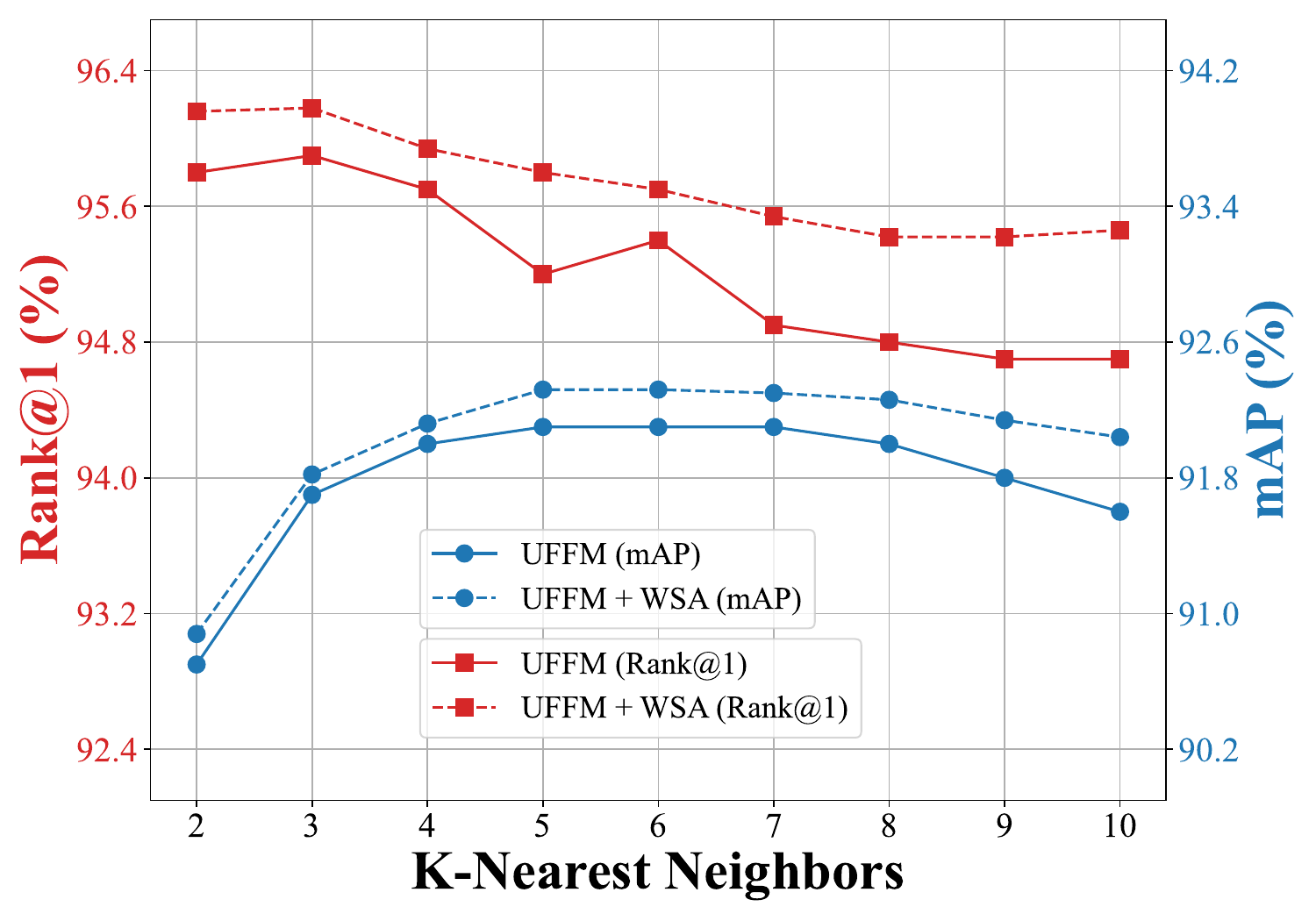}
        \caption{The impact of $K$ on Market-1501 when using our proposed methods}
        \label{impactK}
\end{figure}

\subsection{Parameters analysis} \label{hyper}

\begin{figure}[htp]
  \centering
    \begin{subfigure}{\columnwidth}
        \centering
        \includegraphics[width=0.99\textwidth]{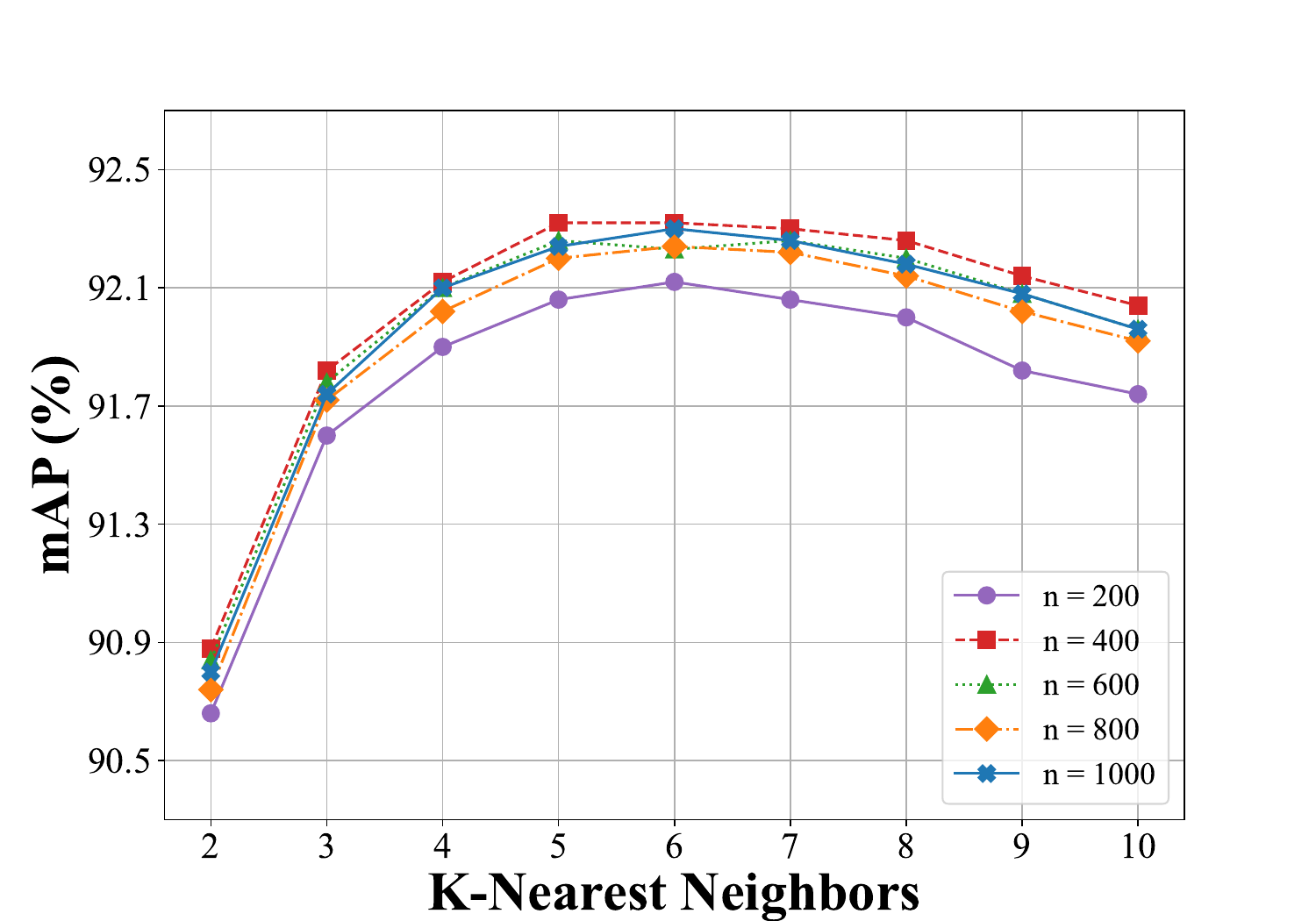}
        \caption{Impact of Parameter $K$ and $n$ on mAP Performance}
    \end{subfigure} 
    \begin{subfigure}{\columnwidth}
        \centering
        \includegraphics[width=0.99\textwidth]{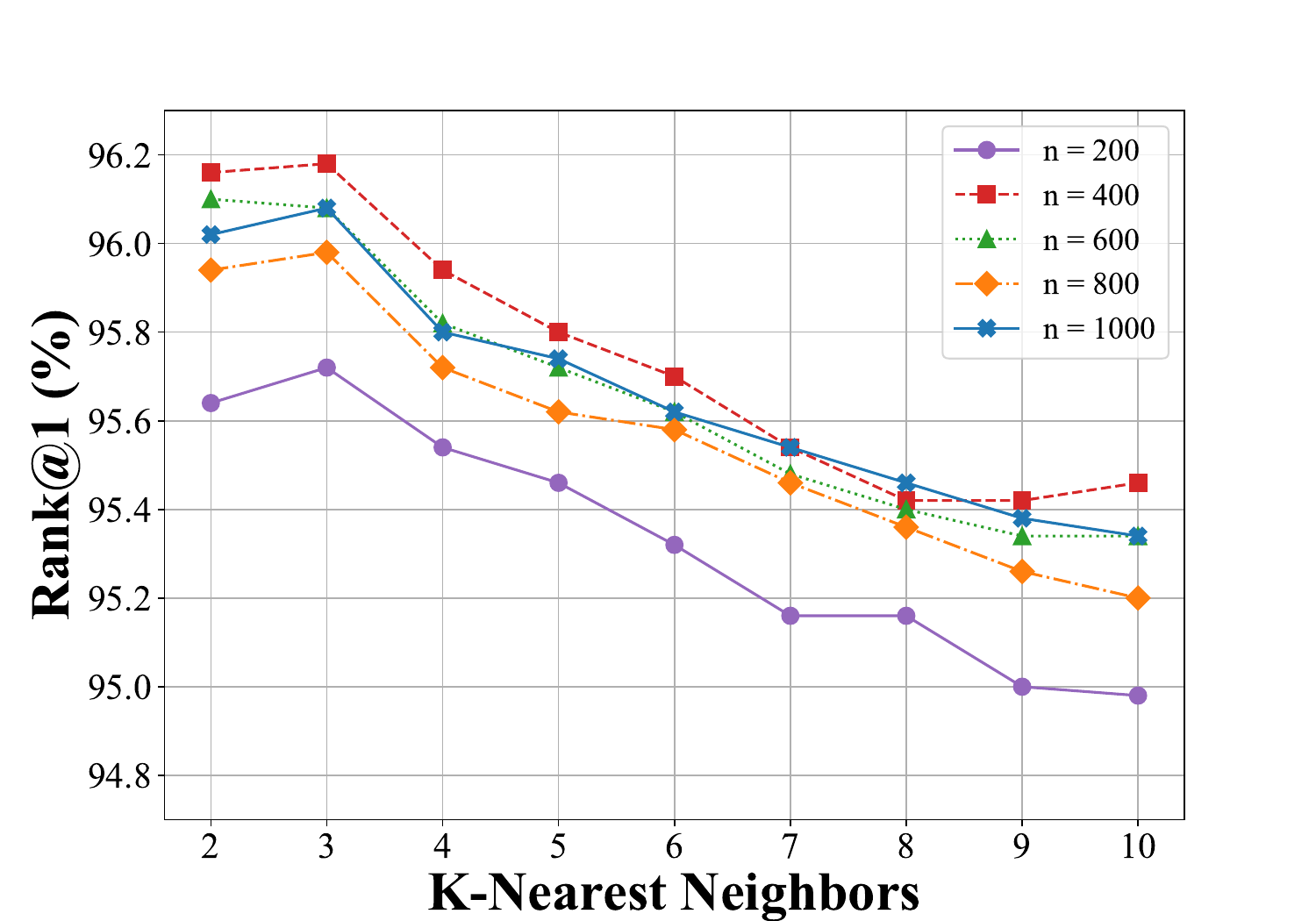}
        \caption{Impact of Parameters $K$ and $n$ on Rank@1 Performance} 
    \end{subfigure} 
    \caption{Impact of hyper parameters $K$ and $n$ on Market-1501}
    \label{impactmarket}
\end{figure}

Our method primarily relies on two parameters: $K$ and $n$, where $K$ directly determines the number of nearest neighbors, and $n$ represents the number of iterations for computing weights in the AMC method. We present the impact of varying these parameters on the Market-1501 dataset. The $K$ coefficient is adjusted from 1 to 10, while $n$ is set to range from 200 to 1000, with incremental steps of 200 for $n$. The metrics in Figure \ref{impactmarket} illustrate that Rank@1 and mAP do not follow a linear trend with increasing values of $K$ or $n$. Notably, with the $K$ coefficient, there is always a trade-off between mAP and Rank@1 when varying $K$. We set $n$ at 400 for both datasets when assessing accuracy changes in both UFFM and UFFM+AMC methods. Our findings, as depicted in Figure \ref{impactK}, indicate that both methods exhibit a trade-off between Rank@1 and mAP with different $K$ values, and the UFFM+AMC method consistently outperforms UFFM across various $K$ coefficients.

\subsection{Combining our method with different backbones.}

In this section, we explore the combination of our method with various backbones, such as Swin Transformer \cite{swin}, ViT-CLIP \cite{clipvit}, Resnet101-iBN \cite{ibn}, and OSNet \cite{osnet}, on Market-1501 and DukeMTMC-ReID datasets. Specifically, we combine our methods with SOLIDER \cite{solider}, CLIP-ReID \cite{clipreid}, RGT$\&$RGPR \cite{RGTRGPR}, and LightMBN \cite{lightMBN} using the backbones Swin Transformer \cite{swin}, ViT-CLIP \cite{clipvit}, Resnet101-iBN \cite{ibn}, and OSNet \cite{osnet}, respectively. The results presented in Table \ref{diffbackbone} demonstrate that the combined performance of UFFM and AMC significantly enhances the performance of different backbones. This confirms that advanced techniques such as UFFM and AMC can be easily integrated with pre-trained models without the need for additional retraining or fine-tuning.

\begin{table}[]
\caption{Performance when combining our method with different backbones.}
\label{diffbackbone}
\centering
\begin{tabular}{lclcc}
\toprule
\multirow{2}{*}{\textbf{Method}} & \multicolumn{2}{c}{\textbf{Market-1501}}           & \multicolumn{2}{c}{\textbf{DukeMTMC-ReID}} \\ \cline{2-5} 
                                 & \textbf{Rank@1} & \multicolumn{1}{c}{\textbf{mAP}} & \textbf{Rank@1}       & \textbf{mAP}       \\ \midrule
SOLIDER \cite{solider}    & 96.9   & 93.9                    & -                & -              \\  \cdashline{2-5} 
+ ours              & \begin{tabular}[c]{@{}c@{}}97.0\\ \textcolor{blue}{$\uparrow$ 0.1}\end{tabular}   & \begin{tabular}[c]{@{}c@{}}94.9\\ \textcolor{blue}{\textbf{$\uparrow$ 1.0}}\end{tabular}                    & -                & -              \\ \midrule
CLIP-ReID \cite{clipreid} & 95.4   & 90.5                    & 90.8             & 83.1           \\ \cdashline{2-5} 
+ ours               & \begin{tabular}[c]{@{}c@{}}96.2\\ \textcolor{blue}{$\uparrow$ 0.8}\end{tabular}   & \begin{tabular}[c]{@{}c@{}}93.3\\ \textcolor{blue}{\textbf{$\uparrow$ 2.8}}\end{tabular}                    & \begin{tabular}[c]{@{}c@{}}92.4\\ \textcolor{blue}{\textbf{$\uparrow$ 1.6}}\end{tabular}              & \begin{tabular}[c]{@{}c@{}}87.2\\ \textcolor{blue}{\textbf{$\uparrow$ 4.1}}\end{tabular}           \\ \midrule
RGT$\&$RGPR \cite{RGTRGPR} & 96.5   & 91.2                    & 92.8             & 84.2           \\  \cdashline{2-5} 
+ ours               & \begin{tabular}[c]{@{}c@{}}96.6\\ \textcolor{blue}{$\uparrow$ 0.1}\end{tabular}   & \begin{tabular}[c]{@{}c@{}}92.8\\ \textcolor{blue}{\textbf{$\uparrow$ 1.6}}\end{tabular}                    & \begin{tabular}[c]{@{}c@{}}93.1\\ \textcolor{blue}{$\uparrow$ 0.3}\end{tabular}             & \begin{tabular}[c]{@{}c@{}}87.5\\ \textcolor{blue}{\textbf{$\uparrow$ 3.3}}\end{tabular}           \\ \midrule
LightMBN \cite{lightMBN}  & 96.3   & 91.5                    & 92.1             & 83.7           \\ \cdashline{2-5} 
+ ours               & \begin{tabular}[c]{@{}c@{}}96.7\\ \textcolor{blue}{$\uparrow$ 0.4}\end{tabular}   & \begin{tabular}[c]{@{}c@{}}94.1\\ \textcolor{blue}{\textbf{$\uparrow$ 2.6}}\end{tabular}                    & \begin{tabular}[c]{@{}c@{}}93.9\\ \textcolor{blue}{\textbf{$\uparrow$ 1.8}}\end{tabular}             & \begin{tabular}[c]{@{}c@{}}88.0\\ \textcolor{blue}{\textbf{$\uparrow$ 4.3}}\end{tabular}           \\ \bottomrule
\end{tabular}
\end{table}

\subsection{Visualization of results}
\begin{figure*}
\centering
\begin{subfigure}{2.1\columnwidth}
        \includegraphics[width=\textwidth]{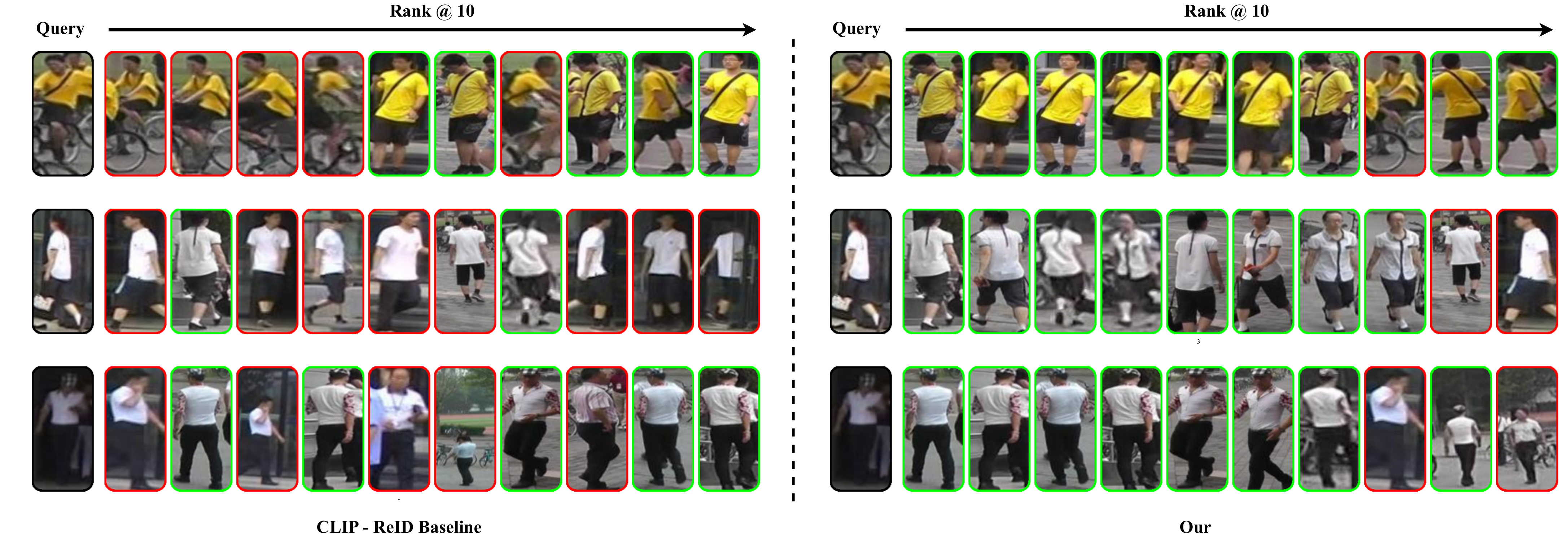}
        \caption{Market1501}
        \label{visualMarket}
\end{subfigure}

\begin{subfigure}{2.1\columnwidth}
        \includegraphics[width=\textwidth]{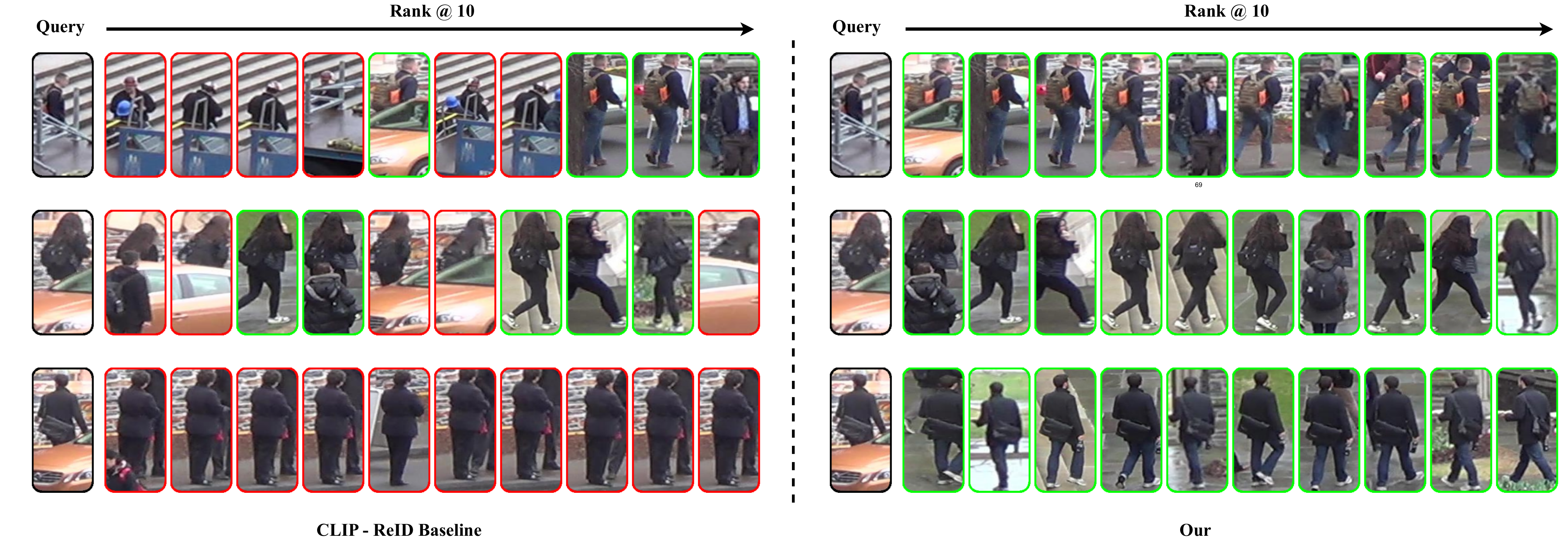}
        \caption{DukeMTMC-ReID} 
        \label{visualDuke}
\end{subfigure}
\caption{\textbf{Visualize results:} The results of the CLIP-ReID Baseline model are on the left, and those of the CLIP-ReID Baseline combined with our method are on the right. Each row displays query images and the top 10 retrieved library images. Green and red boxes indicate true positives and false positives, respectively.}
\label{visual}
\end{figure*}

We present visual examples of the top 10 rankings as determined by the baseline CLIP-ReID model and our proposed method in Figure \ref{visual} for the Market-1501 and DukeMTMC-ReID datasets, respectively. The results demonstrate that the baseline model struggles with query retrieval when subjects change posture or are occluded. Notably, integrating our method significantly enhances performance, effectively handling subjects in various postures within the Market-1501 dataset (Figure \ref{visualMarket}) and achieving robust results even in scenarios where subjects are occluded in the DukeMTMC-ReID dataset (Figure \ref{visualDuke}). These visualizations underscore the efficacy of our approach in Feature Fusion for improved re-identification.

\section{Discussion and Conclusion}\label{conclusion}
\subsection{Limitation}

While our proposed methods effectively leverage multi-view features for person re-identification, there are some limitations to consider. \textit{First}, our methods remain dependent on the hyperparameters $K$ and $n$, where each dataset may have a different set of optimal hyperparameters. To ensure the generalization of the method across different datasets, in this study, we only selected hyperparameters that balance accuracy across various datasets rather than the optimal hyperparameters for each dataset. \textit{Second}, proposed methods require higher computational costs during the querying process compared to using single-view features directly in computing similarity. \textit{Third}, our approach generates multi-view features based on unsupervised selection of neighboring features without needing additional training or fine-tuning of the model. So, the effectiveness of our method depends on the accuracy of pre-trained models.

\subsection{Conclusion}

This study introduces two advanced methodologies: Uncertainty Feature Fusion Method (UFFM) and Auto-weighted Measure Combination (AMC), along with their empirical evaluations on standard datasets such as Market-1501, DukeMTMC-ReID, and MSMT17. These methodologies not only enhance Rank-1 accuracy and mean Average Precision (mAP) but also effectively address challenges posed by image transformations due to varying camera angles and occlusions. The integration of UFFM and ACM provides a robust and versatile approach for person re-identification, paving new paths for the development of surveillance systems, particularly in contexts that demand high accuracy and reliability. We anticipate that these findings will make a significant contribution to the field of camera-based person identification and encourage further research in optimizing and adapting these methods to complex real-world scenarios.


{\small
\bibliography{ref}
}
\end{document}